\documentclass[conference]{IEEEtran}
\IEEEoverridecommandlockouts
\usepackage{cite}
\usepackage{amsmath,amssymb,amsfonts}
\usepackage{algorithmic}
\usepackage{graphicx}
\usepackage{textcomp}
\usepackage{xcolor}
\usepackage{booktabs}
\usepackage{pgfplots}
\usepackage{pgfplotstable}
\usepackage{todonotes}
\usepackage{hyperref}

\pgfplotsset{compat=1.18}

\definecolor{quantumgreen}{RGB}{15,110,83}
\definecolor{classicalgray}{RGB}{95,94,90}

\def\BibTeX{{\rm B\kern-.05em{\sc i\kern-.025em b}\kern-.08em
    T\kern-.1667em\lower.7ex\hbox{E}\kern-.125emX}}

\begin{document}

\title{DynGhost: Temporally-Modelled Transformer for\\
Dynamic Ghost Imagings%
}

\author{
\IEEEauthorblockN{
Vittorio Palladino \qquad
Emadeldeen Hamden \qquad
Ahmet Enis Cetin
}
\thanks{%
V. Palladino is with Politecnico di Milano, Milan, Italy, and the Department of Electrical and Computer Engineering, University of Illinois Chicago, Chicago, IL, USA (e-mail: vpall3@uic.edu).

E. Hamden is with the Department of Electrical and Computer Engineering, University of Illinois Chicago, Chicago, IL, USA (e-mail: ehamda3@uic.edu).

A. E. Cetin is with the Department of Electrical and Computer Engineering, University of Illinois Chicago, Chicago, IL, USA, and with the University of Illinois Urbana--Champaign, Champaign, IL, USA (e-mail: aecyy@uic.edu).}
}

\maketitle

\begin{abstract}
Ghost imaging reconstructs spatial information from a single-pixel bucket
detector by correlating structured illumination patterns with scalar intensity
measurements. While deep learning approaches have achieved promising results
on static scenes, two critical limitations remain unaddressed: existing
architectures fail to exploit temporal coherence across frames, leaving dynamic
ghost imaging largely unsolved, and they assume additive Gaussian noise models
that do not reflect the true Poissonian statistics of real single-photon
hardware. We present \textbf{DynGhost} (Dynamic Ghost Imaging Transformer), a
transformer architecture that addresses both limitations through alternating
spatial and temporal attention blocks. Our detector-aware training framework,
based on physically accurate hardware simulations (SNSPDs, SPADs, SiPMs) and
Anscombe variance-stabilizing normalization, resolves the distribution shift
that causes classical models to fail under realistic hardware constraints.
Experiments across multiple benchmarks demonstrate that DynGhost outperforms
both traditional reconstruction methods and existing deep learning
architectures, with particular gains in dynamic and photon-starved settings.
\end{abstract}

\begin{IEEEkeywords}
ghost imaging, transformer, photon-counting detectors, single-photon detection,
temporal attention, Poisson noise, variance-stabilizing transforms,
dynamic scene reconstruction
\end{IEEEkeywords}

\section{Introduction}
\label{sec:intro}

Ghost imaging is an indirect computational imaging technique that reconstructs spatial information from a single-pixel (bucket) detector by correlating structured illumination patterns with scalar intensity measurements \cite{shapiro2008computational, bromberg2009ghost} got from a single pixel detector. Despite extreme under sampling, sensor noise, and continuous scene motion, spatial information can be computationally recovered from an ensemble of pattern-measurement pairs. The absence of a megapixel focal-plane array makes ghost imaging highly attractive for wavelengths where such arrays are prohibitively expensive, for imaging through strongly scattering media, and in photon-starved regimes \cite{erkmen2010ghost, chandra2023compressive}.

While traditional compressive sensing baselines and recent deep learning methods ranging from convolutional networks \cite{lyu2017deep, wang2019learning} to state-of-the-art transformers like Ghost-GPT \cite{ghostgpt2025} have achieved remarkable reconstruction quality, they suffer from two critical limitations. First, existing architectures treat scenes as purely static, completely failing to exploit temporal coherence across frames. This leaves dynamic ghost imaging essentially unsolved and forces practitioners to either discard moving frames or accept severe motion artefacts. Second, these networks assume additive Gaussian noise, ignoring the true Poissonian statistics, dark counts, and afterpulsing inherent to the single-photon detectors required in real-world applications.

To solve these problems, we propose \textbf{DynGhost} (Dynamic Ghost Imaging Transformer). Our architecture overcomes the static-scene bottleneck by using alternating spatial and temporal attention blocks to propagate motion coherence across frames as an implicit reconstruction prior. To address the detector-model gap, we introduce an end-to-end framework trained and evaluated under physically accurate photon-counting detector simulations, utilizing Anscombe variance-stabilizing normalization to eliminate distribution shifts.

In summary, our main contributions are:
\begin{enumerate}
    \item \textbf{Temporal-aware reconstruction.} The first dynamic ghost imaging transformer that exploits motion coherence via spatial-temporal attention—drawing inspiration from video transformer architectures but uniquely adapted for sequential illumination—and temporal consistency losses, yielding significantly smoother predictions.
    \item \textbf{Detector-aware evaluation and distribution shift mitigation.} The first framework to train and evaluate under true single-photon detector simulations (SNSPDs, SPADs, SiPMs). We identify and resolve the catastrophic distribution shift of Gaussian-trained models, achieving a +33.4\% SSIM gain on real hardware by utilizing correct Poissonian noise modelling and variance-stabilization.
    \item \textbf{Photon-counting deployment characterization.} We benchmark seven
    photon-count normalization strategies, define the operating regime where
    hardware-aware models decisively outperform classical alternatives
    ($\geq$100~photons/measurement, DCR~$<$~10{,}000\,Hz,
    efficiency~$>$~60\%), and provide a structured explanation of failure modes
    under realistic hardware constraints.
\end{enumerate}

The remainder of this paper is organized as follows: Section \ref{sec:prelim} provides the theoretical background on single-photon and classical ghost imaging. Section \ref{sec:method} outlines the problem formulation and details the proposed DynGhost architecture. Sections \ref{sec:experiments} and \ref{sec:conclusion} present our experimental evaluations and concluding remarks.
The code and pretrained models are publicly available at
\url{https://github.com/vittpall/MMSP-26-GhostImaging}.

\section{Preliminaries}
\label{sec:prelim}

To properly contextualize the proposed architecture, it is necessary to explain the operational principles of ghost imaging from its foundational mechanisms to modern computational and deep-learning frameworks, as well as the physical hardware limitations that constrain dynamic imaging.

\subsection{Single-Pixel and Structured Illumination}

Traditional single-pixel imaging techniques often rely on exhaustive raster scanning, which necessitates a sequential capture of $N^2$ independent measurements to fully resolve an $N \times N$ scene. Ghost imaging overcomes this bottleneck by illuminating the unknown object, denoted as $x$, with a sequence of spatially structured light patterns. A bucket detector then records the integrated transmitted or reflected energy, drastically reducing the required number of acquisitions.

The efficiency of this subsampling is expressed by the sampling ratio:
$$\beta = \frac{M}{N^2}$$
where $M$ represents the total number of structured masks projected. For every individual pattern $A^{(m)}$, the corresponding detector response is the scalar inner product of the illumination mask and the object's transmission or reflection profile:
$$b^{(m)} = \sum_{i=1}^{N} \sum_{j=1}^{N} A_{i,j}^{(m)} \cdot x_{i,j}$$
\subsection{Evolution of Reconstruction Paradigms}

Historically, baseline reconstruction techniques like Differential Ghost Imaging (DGI) approximated the scene via weighted averages of illumination masks, resulting in low-fidelity images. Modern computational ghost imaging reframes this as a discrete linear inverse problem ($b = \Psi x$). To achieve sub-Nyquist sampling ($\beta \ll 1$), compressed sensing exploits the inherent sparsity of natural scenes using regularized optimization resolved via iterative solvers like FISTA \cite{beck2009fast} or ADMM \cite{boyd2011distributed}.

In recent years, data-driven methodologies have further pushed the boundaries of image recovery. These neural approaches typically fall into two categories: two-stage refinement—where an initial classical estimate is passed through a network (e.g., U-Net) for denoising and super-resolution—and end-to-end paradigms that map raw bucket sequences directly to the spatial domain (e.g., CNN).

\begin{figure}[htbp]
  \centering
  \includegraphics[width=\linewidth]{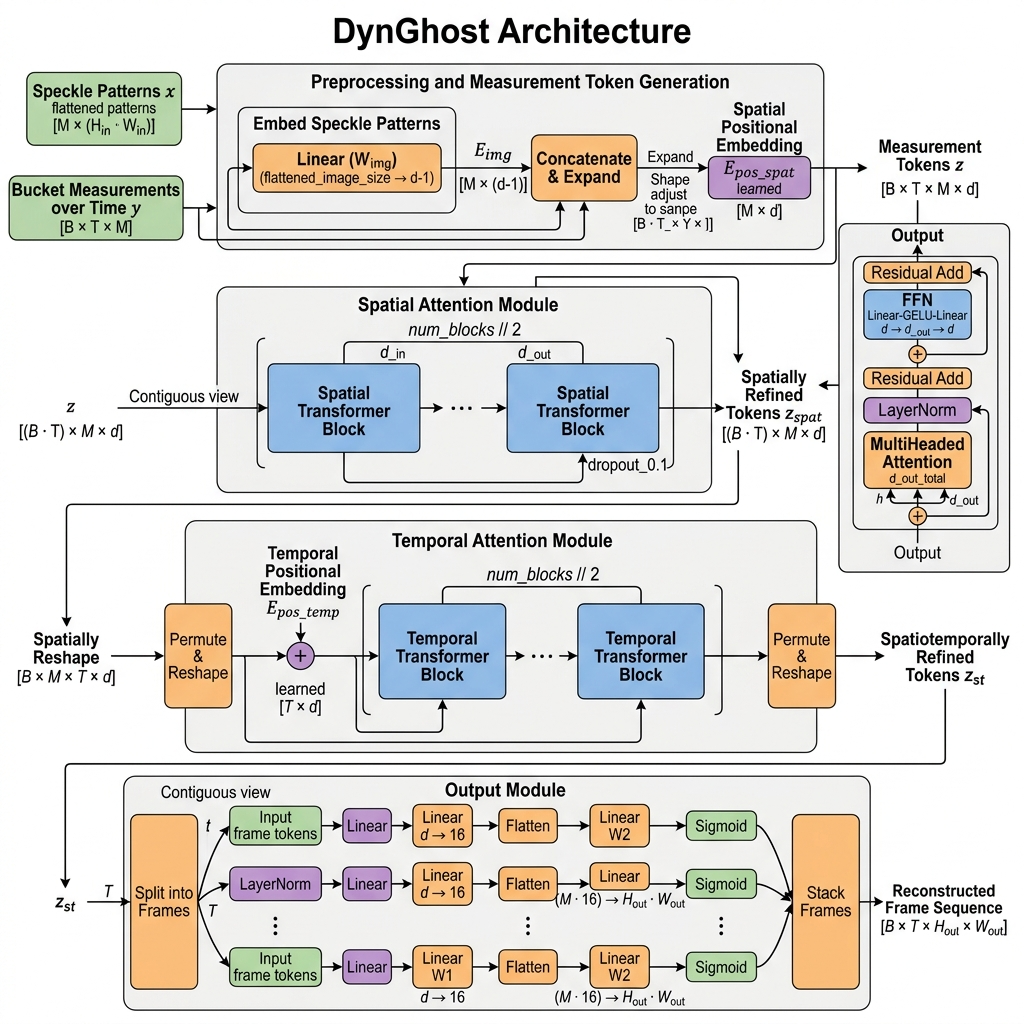}
  \caption{Architecture of Dynghost model.}
  \label{fig:arch}
\end{figure}

\section{Method}
\label{sec:method}

\subsection{Problem Formulation}

A ghost imaging system illuminates an object with $M$ structured speckle
patterns $\{H_i\}_{i=1}^{M}$, $H_i \in \mathbb{R}^{N}$ ($N = W \times
H$). A single-pixel detector returns bucket measurements
$\{b_i\}_{i=1}^{M}$. For a static scene:
$$b_i = \langle H_i, x \rangle + \eta_i$$
where $x\in\mathbb{R}^N$ is the unknown image and $\eta_i$ is
measurement noise. For \textbf{dynamic scenes} we have $T$ frames
$\{x^{(t)}\}_{t=1}^T$ each generating its own measurement sequence
$\{b_i^{(t)}\}_{i=1}^M$, jointly reconstructed to exploit temporal
correlations.

\subsection{Architecture}

Temporal Ghost-GPT processes bucket sequences $B \in \mathbb{R}^{T \times M}$
and outputs frames $\hat{X} \in \mathbb{R}^{T \times H \times W}$.

\textit{Token embedding.} For each frame $t$ and pattern $i$:
\begin{equation}
  z_i^{(t)} = \bigl[\mathrm{Embed}(H_i)\;\|\;b_i^{(t)}\bigr]
  + \mathrm{PE}_{\mathrm{spatial}}(i)
  + \mathrm{PE}_{\mathrm{temporal}}(t),
\end{equation}
where $\mathrm{Embed}(\cdot)$ is a learned linear projection and both
positional embeddings are learned.

\textit{Alternating attention blocks.} The $L$ transformer blocks
alternate between two modes:
\begin{align}
  \text{Spatial:}\quad
    &Z^{(t)} \leftarrow \mathrm{TF}_\mathrm{spatial}(Z^{(t)}),
    \;\forall t, \label{eq:spatial}\\
  \text{Temporal:}\quad
    &Z_i \leftarrow \mathrm{TF}_\mathrm{temporal}(Z_i),
    \;\forall i. \label{eq:temporal}
\end{align}
Spatial blocks model informative patterns per frame; temporal blocks
propagate information across frames to exploit motion coherence.
Output projection is $\hat{x}^{(t)} = \sigma(\mathrm{MLP}(Z^{(t)}))$.
Spatial attention is $\mathcal{O}(M^2)$ per frame; temporal attention
is $\mathcal{O}(T^2)$ per pattern. With $M=188$ and $T=8$, the
temporal overhead is negligible.

\subsection{Training Objective}

The total loss combines reconstruction, perceptual, and temporal
consistency terms:
\begin{align}
  \mathcal{L}_{\mathrm{MSE}}   &= \frac{1}{T} \sum_{t=1}^T \|\hat{x}^{(t)} - x^{(t)}\|_2^2, \\
  \mathcal{L}_{\mathrm{SSIM}}  &= \frac{1}{T} \sum_{t=1}^T \bigl(1 - \mathrm{SSIM}(\hat{x}^{(t)}, x^{(t)})\bigr), \\
  \mathcal{L}_{\mathrm{temp}}  &= \frac{1}{T-1} \sum_{t=1}^{T-1}
    \bigl\|(\hat{x}^{(t+1)}-\hat{x}^{(t)}) - (x^{(t+1)}-x^{(t)})\bigr\|_2^2,\\
  \mathcal{L} &= \mathcal{L}_{\mathrm{MSE}}
    + 0.5\,\mathcal{L}_{\mathrm{SSIM}}
    + 0.1\,\mathcal{L}_{\mathrm{temp}}.
\end{align}

\subsection{Detector Simulation}
\label{sec:detector}

Let $\mu_i^{(t)} = \langle H_i, x^{(t)}\rangle$ be the true normalized
intensity. For a \textbf{classical detector},
$b_i^{(t)} = \mathrm{clip}(\mu_i^{(t)} + \varepsilon)$,
$\varepsilon \sim \mathcal{N}(0,\sigma^2)$. For a \textbf{photon-counting detector}:
\begin{align}
  n_i^{(t)} &\sim \mathrm{Poisson}\!\left(\mu_i^{(t)} \cdot
  \bar{n} \cdot \eta_{\mathrm{eff}}\right), \label{eq:poisson}\\
  d_i^{(t)} &\sim \mathrm{Poisson}(\lambda_d \cdot \Delta t),\\
  b_i^{(t)} &= n_i^{(t)} + d_i^{(t)} + \text{afterpulse} +
  \text{crosstalk},
\end{align}
where $\bar{n}$ is mean photon flux, $\eta_{\mathrm{eff}}$ is detection
efficiency, $\lambda_d$ is dark count rate (DCR), and $\Delta t$ is
integration time. Parameters for each technology are listed in
Table~\ref{tab:detector_params}.

\begin{table}[htbp]
\caption{Detector Simulation Parameters}
\begin{center}
\begin{tabular}{|l|c|c|c|c|}
\hline
\textbf{Parameter} & \textbf{Classical} & \textbf{SNSPD} & \textbf{SPAD} & \textbf{SiPM} \\
\hline
Efficiency $\eta_{\mathrm{eff}}$ & -- & 0.95 & 0.70 & 0.50 \\
\hline
DCR $\lambda_d$ (Hz) & -- & 10 & 1,000 & 100,000 \\
\hline
Dead time (ns) & -- & 40 & 50 & 20 \\
\hline
Afterpulse prob. & -- & 0 & 0.01 & 0.02 \\
\hline
Crosstalk prob. & -- & 0 & 0 & 0.05 \\
\hline
Timing jitter (ps) & -- & 50 & 300 & 100 \\
\hline
Noise model & Gaussian & \multicolumn{3}{c|}{Poisson + above artefacts} \\
\hline
\end{tabular}
\label{tab:detector_params}
\end{center}
\end{table}

We compare seven photon-count normalization strategies and find that
variance-stabilizing transforms the Anscombe transform
$f(n) = 2\sqrt{n + 3/8}$ and the Freeman--Tukey transform
$f(n) = \sqrt{n} + \sqrt{n+1}$ dramatically outperform all
alternatives by rendering Poisson noise approximately Gaussian with unit
variance, matching the implicit assumption of MSE loss.

\subsection{Experimental Setup}
\label{sec:experiments}

\textit{Datasets.} We evaluate our proposed architecture across three distinct benchmarks: (1) \textbf{Moving MNIST Ghost Imaging} \cite{lecun1998gradient}, featuring digits animated with six unique trajectory types at varying velocities (1--50\,px/frame). Sequences consist of $T=8$ frames at $256\times256$ resolution, partitioned into 5,000 training and 500 validation sequences. (2) \textbf{KViSAR}, a real-world infrared video dataset utilized to test the model's capabilities on non-synthetic, unstructured motion.

\textit{Measurement Model.} Illumination is simulated using $M=188$ structured speckle patterns derived from a physical dual-comb ghost imaging system \cite{ghostgpt2025}. This corresponds to a highly constrained sub-Nyquist sampling ratio of $\beta = 0.29\%$.

\textit{Baselines.} We benchmark DynGhost against established classical algorithms, including Differential Ghost Imaging (DGI) \cite{ferri2010differential}, Pseudo-Inverse (PI), and FISTA (200 iterations) \cite{beck2009fast}. For deep learning baselines, we compare against a standard U-Net \cite{ronneberger2015unet} (applied independently per frame), a CNN adapted from Lyu et al. \cite{lyu2017deep}, and the static Ghost-GPT \cite{ghostgpt2025} model. All neural baselines were trained on the identical dataset and loss formulation to guarantee a fair comparison.

\textit{Implementation Details.} The DynGhost architecture utilizes 8 alternating transformer blocks (4 spatial, 4 temporal) with 8 attention heads and an embedding dimension of 32, totaling approximately 270M parameters. Training was conducted using the AdamW optimizer \cite{loshchilov2019decoupled} (learning rate $3\times10^{-4}$, weight decay $10^{-3}$) with a batch size of 4 over 30 epochs on a single NVIDIA A100 GPU. We evaluate two distinct variants: the \textit{base model} (trained on standard Gaussian noise) and the \textit{detector-aware model} (trained using SNSPD physical simulations and Anscombe normalization).

\subsection{Dynamic Scene Reconstruction}

As detailed in Table~\ref{tab:main_results}, DynGhost significantly outperforms both classical and contemporary deep-learning baselines. Compared to the static Ghost-GPT \cite{ghostgpt2025}, our model achieves a 45\% reduction in MSE and a 16\% improvement in SSIM, despite the added complexity of dynamic motion on Moving MNIST. Furthermore, it delivers a $42\times$ reduction in MSE compared to DGI and operates $130\times$ faster than iterative solvers like FISTA. The per-frame latency of 8.1\,ms falls well below standard real-time video thresholds, highlighting the computational efficiency of batched temporal attention.

To thoroughly assess the  capabilities of our proposed architecture, we extend our
evaluation across one other datasets presenting varying degrees of spatial complexity and
motion unpredictability. As shown in Table~\ref{tab:main_results}, DynGhost consistently
maintains strong performance across the real-world Kvasir
Endoscopy dataset, outperforming all classical and convolutional baselines by a wide margin.

On the Kvasir Endoscopy dataset, GhostGPT achieves a marginal improvement in both SSIM
($0.6222$ vs.\ $0.6112$) and MSE ($0.0133$ vs.\ $0.0140$). We attribute this to a
fundamental architectural trade-off: DynGhost's temporal attention blocks are calibrated to
exploit inter-frame motion coherence, a prior that is abundant in structured synthetic
sequences (Moving MNIST) but significantly weaker in endoscopic video, where
scene-level motion is slow and dominated by fine-grained local texture variation rather than
rigid macroscopic object displacement. In this regime, GhostGPT's purely spatial attention operating
on a richer single-frame parameter budget without the overhead of cross-frame aggregation can match or
marginally surpass our temporal prior. This highlights that for scenes dominated by non-rigid, microscopic texture variations, a purely spatial approach remains highly competitive. Crucially, both methods vastly outperform all
non-transformer baselines (CNN, U-Net, DGI, PI, FISTA) on every dataset and metric, confirming
that the competitive gap is confined to the top two transformer architectures and does not
undermine DynGhost's broader contribution.

\begin{table}[htbp]
\caption{Comprehensive Performance Comparison Across Datasets. Inference times refer to
processing on a single NVIDIA A100 GPU. Because DynGhost processes spatial and temporal data
simultaneously, its reported time is the average per-frame latency (derived by dividing its
full 8-frame sequence processing time by 8) to ensure a fair comparison.}
\begin{center}
\resizebox{\columnwidth}{!}{%
\begin{tabular}{|l|l|c|c|c|}
\hline
\textbf{Dataset} & \textbf{Method} & \textbf{SSIM} $\uparrow$ & \textbf{MSE} $\downarrow$ & \textbf{Time (ms)} $\downarrow$ \\
\hline
\textbf{Moving MNIST}
& \textbf{DynGhost (ours)} & $\mathbf{0.9167 \pm 0.0229}$ & $\mathbf{0.0043 \pm 0.0027}$ & \textbf{7.3} \\
& GhostGPT               & $0.7880 \pm 0.0760$          & $0.0080 \pm 0.0090$          & 14.0 \\
& U-Net                  & $0.8286 \pm 0.0704$          & $0.0226 \pm 0.0162$          & 13.2 \\
& CNN                    & $0.8248 \pm 0.0647$          & $0.0249 \pm 0.0158$          & 8.9  \\
& FISTA                  & $0.3655 \pm 0.0436$          & $0.0425 \pm 0.0198$          & 37,429.6 \\
& PI                     & $0.0769 \pm 0.0156$          & $0.0830 \pm 0.0185$          & 16,760.8 \\
& DGI                    & $0.0579 \pm 0.0156$          & $0.1971 \pm 0.0374$          & 83.2 \\
\hline
\textbf{Kvasir Endoscopy}
& GhostGPT               & $\mathbf{0.6222 \pm 0.0872}$ & $\mathbf{0.0133 \pm 0.0081}$ & 221.6 \\
& \textbf{DynGhost (ours)} & $0.6112 \pm 0.0822$        & $0.0140 \pm 0.0068$          & \textbf{166.6} \\
& CNN                    & $0.5655 \pm 0.0743$          & $0.0252 \pm 0.0097$          & 4.2  \\
& U-Net                  & $0.4341 \pm 0.0628$          & $0.0553 \pm 0.0172$          & 8.2  \\
& DGI                    & $0.3057 \pm 0.0465$          & $0.0907 \pm 0.0174$          & 84.9 \\
& PI                     & $0.2685 \pm 0.0553$          & $0.0477 \pm 0.0127$          & 15,825.0 \\
& FISTA                  & $0.2493 \pm 0.0302$          & $0.0770 \pm 0.0228$          & 40,346.3 \\
\hline
\end{tabular}%
}
\label{tab:main_results}
\end{center}
\end{table}

\subsection{Ablation Studies}

\subsubsection{Contributions of architecture}
To understand the specific contributions of our architectural choices, we run targeted ablations (Table~\ref{tab:ablation}) alongside an analysis of how sequence length, motion type, and absolute object velocity impact structural fidelity.

\begin{table}[htbp]
\caption{Component Ablation (10 Epochs, 2,000 Samples)}
\begin{center}
\begin{tabular}{|l|c|c|c|}
\hline
\textbf{Variant} & \textbf{MSE} $\downarrow$ & \textbf{SSIM} $\uparrow$ & \textbf{T. Cons.} $\downarrow$ \\
\hline
Full model & \textbf{0.0044} & \textbf{0.917} & \textbf{0.012} \\
\hline
No temporal attention & 0.0101 & 0.889 & 0.025 \\
\hline
No temporal pos.\ enc. & 0.0055 & 0.890 & 0.015 \\
\hline
MSE loss only & 0.0035 & 0.670 & 0.013 \\
\hline
No temp.\ consistency loss & 0.0048 & 0.910 & 0.018 \\
\hline
1 temporal block (vs.\ 4) & 0.0060 & 0.870 & 0.016 \\
\hline
\end{tabular}
\label{tab:ablation}
\end{center}
\end{table}

Removing the temporal attention blocks increases the MSE by a factor of 2.3, confirming that cross-frame feature aggregation is the primary driver of performance in dynamic scenes. Similarly, relying solely on an MSE loss function causes a severe collapse in SSIM (from 0.917 to 0.670), demonstrating the absolute necessity of perceptual loss components for high-fidelity imaging.

\begin{figure}[htbp]
  \centering
  \includegraphics[width=\linewidth]{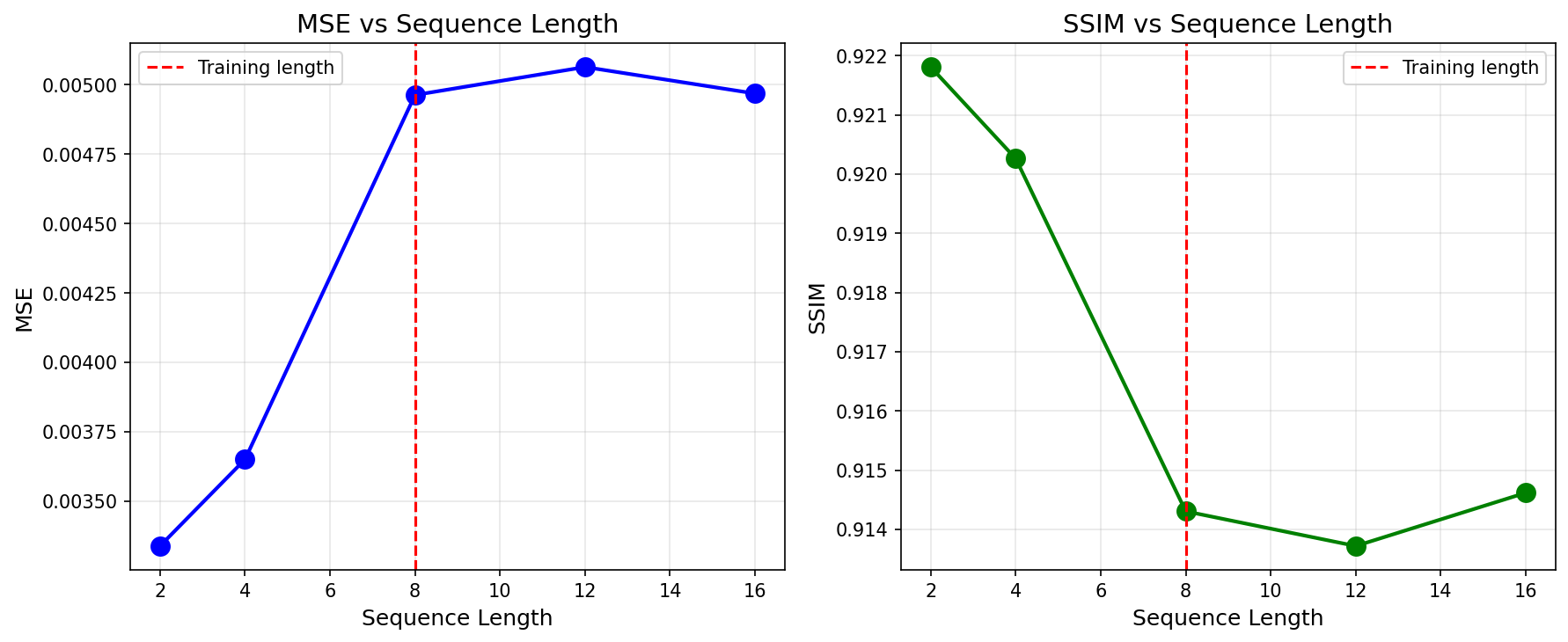}
  \caption{Ablation on sequence length ($T$). Performance peaks at the training length ($T=8$). Shorter sequences lack sufficient temporal context to resolve spatial ambiguities, while over-extending the sequence length introduces compounding motion tracking errors.}
  \label{fig:seq_length}
\end{figure}

\begin{figure}[htbp]
  \centering
  \includegraphics[width=\linewidth]{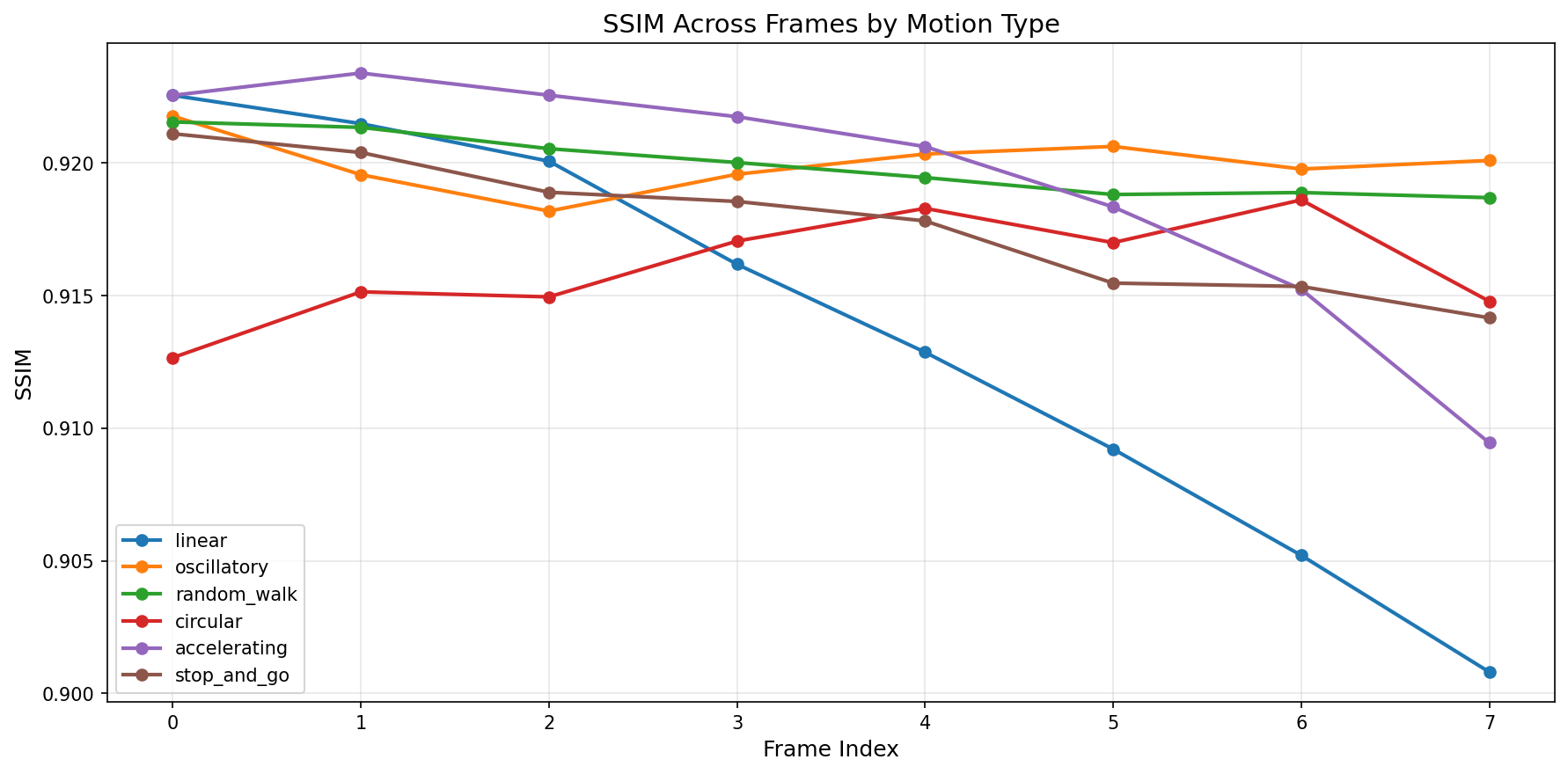}
  \caption{Frame-by-frame SSIM degradation segmented by motion type. While predictable trajectories (linear, oscillatory) maintain stable fidelity, highly dynamic or erratic motions (random walk, accelerating) exhibit a noticeable decay in SSIM across later frames as temporal correlation weakens.}
  \label{fig:frame_ssim_motion}
\end{figure}

\begin{figure}[htbp]
  \centering
  \includegraphics[width=\linewidth]{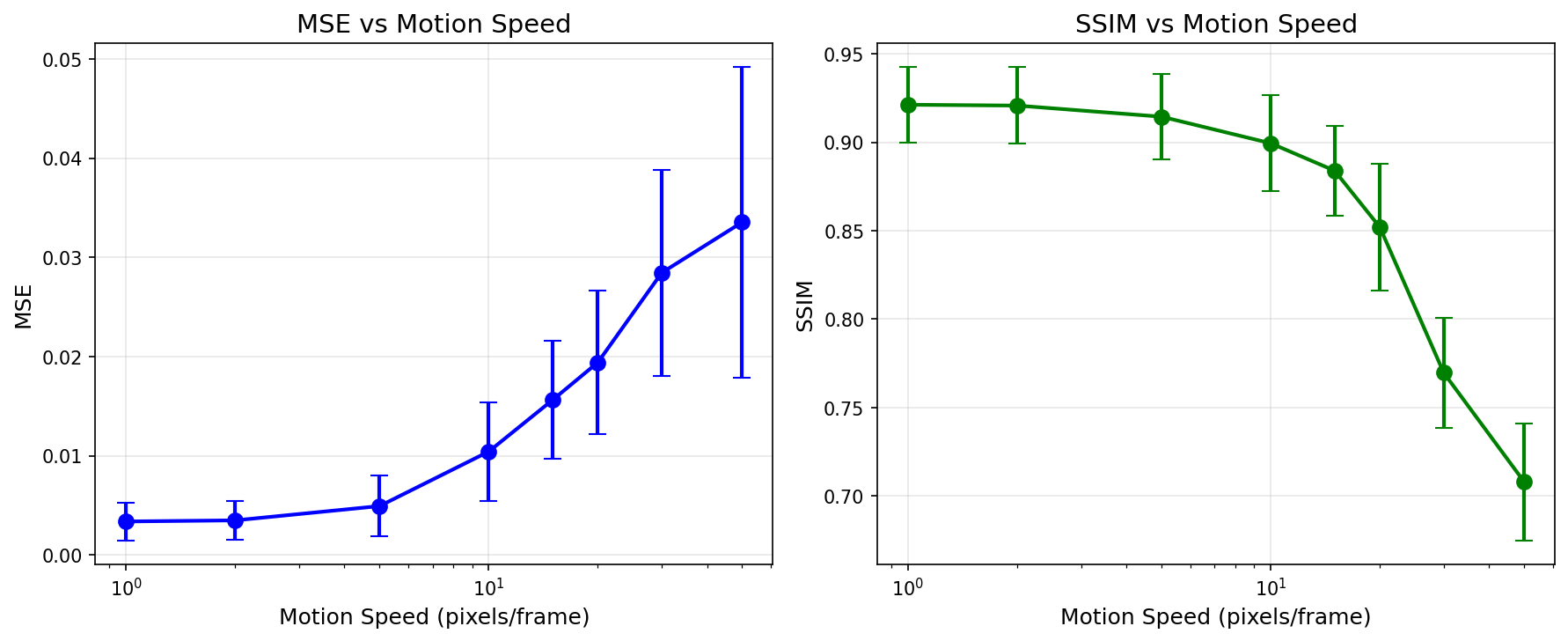}
  \caption{Quantitative evaluation of reconstruction fidelity across varying object velocities (log scale). Error bars indicate standard deviation. The model is highly resilient at low-to-moderate speeds (up to 10\,px/frame), with degradation accelerating at extreme velocities as intra-frame motion blur begins to dominate the measurement period.}
  \label{fig:motion_speed}
\end{figure}

\subsubsection{Robustness across varied kinematics}
As shown in Figures~\ref{fig:frame_ssim_motion} and \ref{fig:motion_speed},
DynGhost remains stable for linear and oscillatory motions, while erratic
trajectories (random walks, sudden accelerations) cause a gradual SSIM decline
toward the end of the sequence. Velocity-wise, SSIM stays above $0.90$ up to
10\,px/frame and above $0.85$ up to 20\,px/frame. Degradation at extreme
speeds (SSIM~$\approx 0.70$ at 50\,px/frame) is physically expected: intra-frame
motion blur at such velocities erases high-frequency spatial detail before the
bucket detector can integrate it.

\subsubsection{Sampling ratio}

We further ablate the number of structured illumination patterns $M$, which directly controls the information available per frame. As reported in Table~\ref{tab:sampling_ablation}, DynGhost is evaluated on Moving MNIST at five operating points spanning $M \in \{94, 188, 376, 752\}$ patterns over a $256\times256$ scene ($N^2 = 65{,}536$ pixels).

\begin{table}[htbp]
\caption{Sampling Ratio Ablation on Moving MNIST (full 30-epoch training)}
\begin{center}
\begin{tabular}{|c|c|c|c|}
\hline
\textbf{Patterns $M$} & \textbf{Ratio $\beta$} & \textbf{SSIM} $\uparrow$ & \textbf{MSE} $\downarrow$ \\
\hline
94  & 0.14\% & $0.8722 \pm 0.0383$ & 0.01040 \\
\hline
188 & 0.29\% & $0.9167 \pm 0.0229$ & 0.00430 \\
\hline
376 & 0.57\% & $\mathbf{0.9190 \pm 0.0212}$ & 0.00388 \\
\hline
752 & 1.15\% & $0.9188 \pm 0.0219$ & $\mathbf{0.00374}$ \\
\hline
\end{tabular}
\label{tab:sampling_ablation}
\end{center}
\end{table}

The results reveal two distinct regimes. In the low-measurement regime ($M \leq 188$), reconstruction quality improves sharply with each doubling of patterns: going from $M=94$ to $M=188$ reduces MSE by $52\%$ and lifts SSIM by $+3.6$ points. This confirms that below $\beta \approx 0.3\%$, the bucket sequence is severely information-limited and additional patterns directly resolve spatial ambiguities that temporal attention alone cannot compensate for. In the high-measurement regime ($M \geq 376$), gains saturate rapidly: doubling patterns from $376$ to $752$ yields only a marginal MSE reduction with a negligible SSIM change, indicating that the temporal prior has absorbed most of the residual reconstruction uncertainty.

\subsection{Noise Robustness and Missing Measurements}
\label{sec:noise_robustness}

In practical low-dose or high-speed deployments, ghost imaging systems are frequently subject to severe signal-to-noise ratio (SNR) degradation and packet loss. To evaluate this, we subjected the algorithms to varying levels of additive noise. 

\begin{figure}[htbp]
  \centering
  \includegraphics[width=\linewidth]{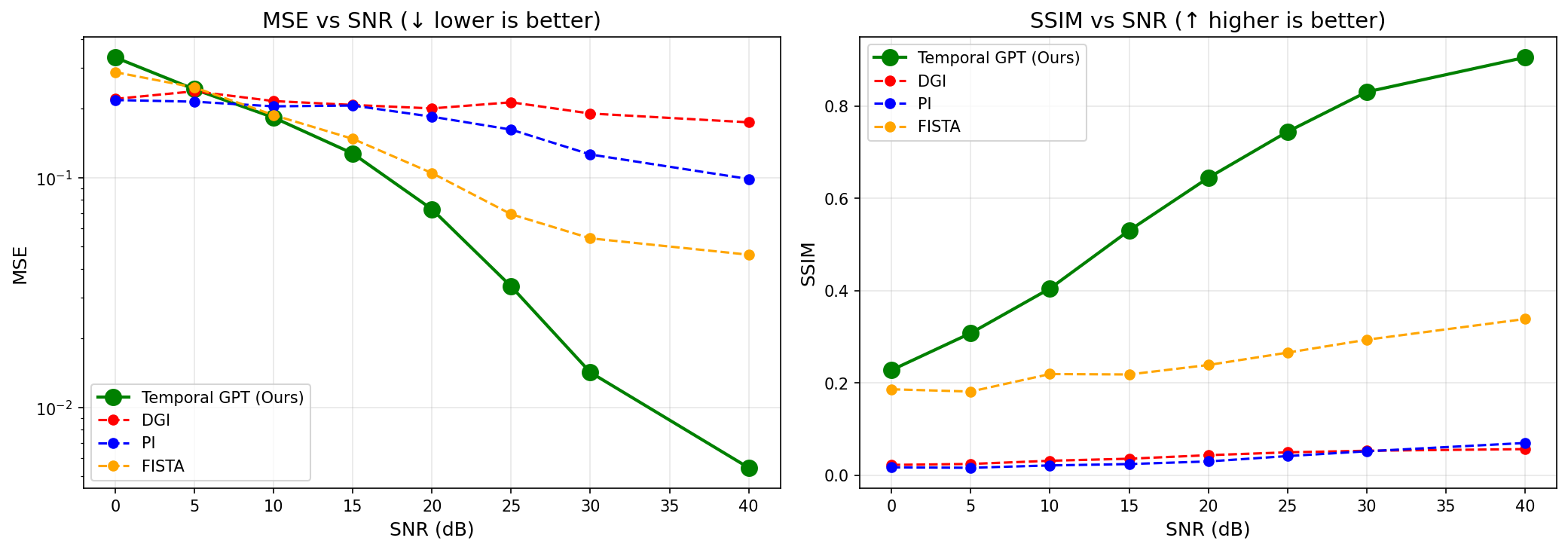}
  \caption{Quantitative noise robustness. DynGhost (Temporal GPT) maintains superior MSE and SSIM across a wide range of Input SNRs compared to standard reconstruction algorithms (DGI, PI, and FISTA), opening a massive performance gap at standard operational levels (15--30 dB).}
  \label{fig:snr_analysis}
\end{figure}

\begin{figure}[htbp]
  \centering
  \includegraphics[width=\linewidth]{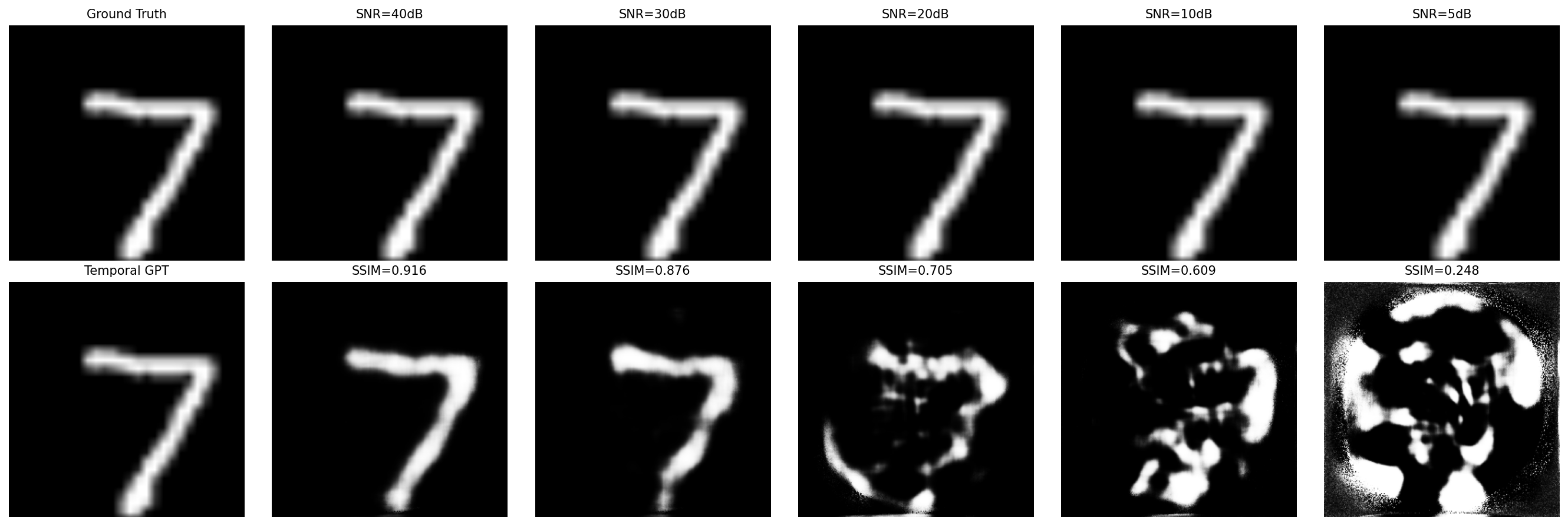}
  \caption{Qualitative reconstruction outputs across varying SNR levels. The proposed model retains strong structural integrity and readability down to 10~dB. Below 10~dB, visual features degrade, though it remains significantly more coherent than classical baselines, which fail to reconstruct entirely at these limits.}
  \label{fig:snr_reconstructions}
\end{figure}

As illustrated in Figure \ref{fig:snr_analysis}, DynGhost dominates classical solvers across the entire noise spectrum. The model maintains an acceptable SSIM ($>0.5$) down to highly degraded 15\,dB SNR environments. Visual evidence of this resilience is provided in Figure \ref{fig:snr_reconstructions}; while the model smoothly blurs at 10\,dB and 5\,dB, it consistently avoids the catastrophic high-frequency noise collapse characteristic of Pseudo-Inverse and FISTA methods.

\subsection{Photon-Counting Detector Evaluation}
\label{sec:quantum}

Although ghost imaging often utilizes single-photon counting detectors (governed by Poisson statistics), existing deep learning approaches ubiquitously train using Gaussian noise models. From a machine learning perspective, testing a Gaussian-trained model on Poisson-distributed hardware data introduces a severe, yet physically predictable, distribution shift.

\subsubsection{Detector Type Comparison}

\begin{table}[htbp]
\caption{SSIM and MSE by Detector Type at 100 Mean Photons}
\begin{center}
\begin{tabular}{|l|c|c|c|c|}
\hline
 & \multicolumn{2}{c|}{\textbf{Base (Gaussian)}} & \multicolumn{2}{c|}{\textbf{Detector-aware (Poisson)}} \\
\cline{2-5}
\textbf{Detector} & \textbf{\textit{MSE}} $\downarrow$ & \textbf{\textit{SSIM}} $\uparrow$ & \textbf{\textit{MSE}} $\downarrow$ & \textbf{\textit{SSIM}} $\uparrow$ \\
\hline
Classical & 0.0868 & 0.616 & 0.0904 & 0.734 \\
\hline
SNSPD & 0.0871 & 0.628 & \textbf{0.0195} & \textbf{0.838} \\
\hline
SPAD & 0.0869 & 0.627 & 0.0243 & 0.815 \\
\hline
SiPM & 0.0874 & 0.636 & 0.0889 & 0.730 \\
\hline
\end{tabular}
\label{tab:detector_comparison}
\end{center}
\end{table}

The base Gaussian model achieves nearly identical SSIM across all detectors (0.62--0.64), confirming its inability to adapt to specific hardware artifacts. The detector-aware model, however, breaks this symmetry: SNSPD and SPAD reconstructions improve by +34\% and +30\%, respectively. The Silicon Photomultiplier (SiPM) shows no improvement due to its excessive dark count rate (DCR) of $10^5$\,Hz, which buries the true signal, driving the signal-to-dark ratio below 1.

\subsubsection{Normalization Strategy}

\begin{table}[htbp]
\caption{Normalization Strategy Comparison (SNSPD, 100 Mean Photons)}
\begin{center}
\begin{tabular}{|l|c|c|}
\hline
\textbf{Method} & \textbf{MSE} $\downarrow$ & \textbf{SSIM} $\uparrow$ \\
\hline
None & 0.0605 & 0.725 \\
\hline
$\sqrt{\cdot}$ & 0.0846 & 0.752 \\
\hline
$\log(1+\cdot)$ & 0.0852 & 0.750 \\
\hline
Min-max & 0.0851 & 0.741 \\
\hline
Z-score & 0.0846 & 0.739 \\
\hline
Anscombe $2\sqrt{n+3/8}$ & 0.0192 & 0.842 \\
\hline
Freeman--Tukey $\sqrt{n}+\sqrt{n+1}$ & \textbf{0.0191} & \textbf{0.844} \\
\hline
\end{tabular}
\label{tab:normalisation}
\end{center}
\end{table}

Standard min-max and Z-score normalizations fail on single-photon data because they treat noise as a uniform scaling problem. However, Poisson noise is fundamentally heteroscedastic (variance scales with the mean). Variance-stabilizing transforms-specifically Anscombe and Freeman-Tukey convert Poisson counts to approximately unit-variance Gaussian variables. As shown in Table \ref{tab:normalisation}, this aligns the input distribution with the implicit assumptions of the MSE loss function, drastically improving SSIM.

\section{Discussion}
\label{sec:discussion}

The integration of temporal attention acts as a powerful learned regularizer. By sharing information across frames, the model actively reduces reconstruction ambiguity at every individual time step a process analogous to how modern video codecs exploit inter-frame redundancy. The resulting $2.3\times$ MSE improvement validates that temporal coherence is a massive, largely untapped prior in computational ghost imaging.

Furthermore, our hardware-aware detector evaluations reveal a critical insight into normalization bottlenecks. A significant limitation of our current framework is the fixed photon flux assumption. The variance-stabilizing normalizers (Anscombe and Freeman-Tukey) are calibrated at a specific photon count ($N=100$) and their effectiveness demonstrably degrades under variable illumination (flux shift). This strictly limits the immediate practical utility of the approach to laboratory or highly controlled illumination environments. To deploy this reliably in unconstrained real-world settings, future deployments must incorporate adaptive normalization layers capable of estimating Anscombe parameters dynamically based on real-time moving averages of the flux. Similarly, our hardware profiling indicates that a detector's Dark Count Rate is a far greater obstacle than its base efficiency.

\textit{Limitations.} This work has several limitations. The quadratic
scaling of temporal attention $\mathcal{O}(T^2)$ restricts the model to
short sequences ($T=8$), and the hardware-aware normalization is calibrated
at a fixed photon flux, degrading under variable illumination as discussed. Furthermore, validation on physical real-time dynamic ghost imaging hardware remains an
important open step.

\textit{Future directions} include scaling the model to handle natural medical or biological imagery better (e.g., fluorescence microscopy), integrating adaptive flux normalization, and exploring state-space alternatives like Mamba \cite{gu2023mamba} to handle significantly longer temporal sequence lengths without quadratic attention overheads.

\section{Conclusion}
\label{sec:conclusion}

We have presented DynGhost, the first transformer-based architecture natively designed for dynamic ghost imaging, coupled with a systematic integration of single-photon detector physics. By leveraging temporal attention, DynGhost yields a 45\% reduction in MSE and a 16\% increase in SSIM over static benchmarks. Furthermore, utilizing detector-aware variance-stabilizing normalization adds an additional +33\% SSIM over naively trained models. We have identified a definitive operating regime (100 photons/measurement, DCR $<$ 10,000\,Hz, efficiency $>$ 60\%) where hardware-aware models decisively outperform classical alternatives. These findings provide a robust, principled blueprint for deploying high-speed ghost imaging in photon-starved environments.



\bibliographystyle{IEEEtran}
\bibliography{references}

\end{document}